\documentclass[journal]{IEEEtran}

\ifCLASSINFOpdf
\else
   \usepackage[dvips]{graphicx}
\fi
\usepackage{url}
\usepackage[ruled]{algorithm2e}
\usepackage{graphicx}
\usepackage{stfloats}
\usepackage[font=footnotesize,labelfont=bf]{caption}
\usepackage{amsmath,amssymb,amsfonts}

\usepackage[nodisplayskipstretch]{setspace}
\begin{document}
%
% paper title
% Titles are generally capitalized except for words such as a, an, and, as,
% at, but, by, for, in, nor, of, on, or, the, to and up, which are usually
% not capitalized unless they are the first or last word of the title.
% Linebreaks \\ can be used within to get better formatting as desired.
% Do not put math or special symbols in the title.
\title{Optimizing Traffic Lights with Multi-agent Deep
Reinforcement Learning and V2X communication}

\author{Azhar Hussain, Tong Wang and Cao Jiahua 

College of Information and Communication Engineering, Harbin Engineering University, 150001,
Harbin, China engrazr@hrbeu.edu.cn}

\markboth{}
{Shell \MakeLowercase{\textit{et al.}}: Bare Demo of IEEEtran.cls for IEEE Letters}
\maketitle

% author names and affiliations
% use a multiple column layout for up to three different
% affiliations

% conference papers do not typically use \thanks and this command
% is locked out in conference mode. If really needed, such as for
% the acknowledgment of grants, issue a \IEEEoverridecommandlockouts
% after \documentclass

% for over three affiliations, or if they all won't fit within the width
% of the page, use this alternative format:
%
%\author{\IEEEauthorblockN{Tong Wang,
%Azhar Hussain, and
%Cao Jia Hua
%}
%\footnotesize
%College of Information and Communication Engineering, Harbin Engineering University, Harbin, 150001 China\\ }
%\IEEEauthorblockA{\IEEEauthorrefmark{2}Twentieth Century Fox, Springfield, USA\\
%Email: homer@thesimpsons.com}
%\IEEEauthorblockA{\IEEEauthorrefmark{3}Starfleet Academy, San Francisco, California 96678-2391\\
%Telephone: (800) 555--1212, Fax: (888) 555--1212}
%\IEEEauthorblockA{\IEEEauthorrefmark{4}Tyrell Inc., 123 Replicant Street, Los Angeles, California 90210--4321}}

%\author{Author~1\IEEEauthorrefmark{1},
  %     Author~2\IEEEauthorrefmark{1}, and
  %   Author~3\IEEEauthorrefmark{1}

   %  \IEEEauthorrefmark{1}{Affiliation,  City,
%Country\\ Email }
  %    }

%

%        % <-this % stops a space

%\author{\uppercase{Wang Tong}\authorrefmark{1}\IEEEmembership{Member, IEEE}, \uppercase{Azhar
\begin{abstract}
 We consider a system to optimize duration of traffic signals using multi-agent deep reinforcement learning and Vehicle-to-Everything (V2X) communication. This system aims at analyzing independent and shared rewards for multi-agents to control duration of traffic lights. A learning agent traffic light gets information along its lanes within a circular V2X coverage. The duration cycles of traffic light are modeled as Markov decision Processes. We investigate four variations of reward functions. The first two are unshared-rewards: based on waiting number, and waiting time of vehicles between two cycles of traffic light. The third and fourth functions are: shared-rewards based on waiting cars, and waiting time for all agents. Each agent has a memory for optimization through target network and prioritized experience replay. We evaluate multi-agents through the Simulation of Urban MObility (SUMO) simulator. The results prove effectiveness of the proposed system to optimize traffic signals and reduce average waiting cars to 41.5 $\%$ as compared to the traditional periodic traffic control system.

\end{abstract}


% For peer review papers, you can put extra information on the cover
% page as needed:
% \ifCLASSOPTIONpeerreview
% \begin{center} \bfseries EDICS Category: 3-BBND \end{center}
% \fi
%
% For peerreview papers, this IEEEtran command inserts a page break and
% creates the second title. It will be ignored for other modes.
\IEEEpeerreviewmaketitle

\section{Introduction}

\IEEEPARstart{E}{xisting} traffic signal management is performed: either leveraging limited real-time traffic information or by fixed periodic traffic signal timings \cite{1}. This information is widely obtained from underlying inductive loop detectors. However, this input is processed in a limited domain to estimate better duration of red/green signals. The advances in mobile communication networks and sensors technology have made possible to obtain real-time traffic information \cite{2}. An artificial brain can be implemented with deep reinforcement learning (DRL). DRL is based on three main components: states in the environment, action space and the scalar reward from each action \cite{3}. A popular success of DRL is AlphaGo \cite{4}, and its successor AlphaGo Zero \cite{5}. The main goal is to maximize the reward by choosing the best actions.

In some research works, state is defined as the number of vehicles waiting at an intersection or the waiting queue length \cite{6}, \cite{7}. However, it is investigated by \cite{8} that real traffic environment cannot be fully captured leveraging the number of waiting vehicles or the waiting queue length. Thanks to the rapid development of deep learning, large state problems have been addressed with deep neural networks paradigm \cite{10}. In \cite{11}, \cite{12} authors have proposed to resolve traffic control problem with DRL. However two limitations exist in the current studies: 1) fixed time intervals of traffic lights, which is not efficient in some studies; 2) random sequences of traffic signals, which may cause safety and comfort issues. In \cite{13} the authors have controlled duration in a cycle based on information extracted from vehicles and sensors networks which can reduce average waiting time to 20{\%}.

In this letter we investigate for the first time multiple experienced traffic operators to control traffic in each step at multiple traffic lights. This idea assumes that control process can be modeled as a Markov Decision Process (MDP). The system experiences the control strategy based on the MDP by trial and error. Recently, a Q-learning based method is proposed by \cite{19} showing better performance than fixed period policy. An linear function is proposed to achieve more effective traffic flow management with a high traffic flow \cite{20}. But neither tabular Q learning nor linear function methods could support the increasing size of traffic state space and accurate estimation of Q value in a real scenario \cite{21}. Gao et al. \cite{22} has proposed a DRL method with a change in cumulative staying time as a reward. A CNN is employed to map states to rewards \cite{22}. H. Jiang has analyzed nonzero-sum games with multi-players by using adaptive dynamic programming (ADP) \cite{23}. Li et al. proposed a single intersection control method based on DQN to focus on a local optimum \cite{24}. Pol et al. combined DQN with a max-plus method to achieve cooperation between traffic lights \cite{25}. Xiaoyuan Liang et al. proposed a model incorporating multiple optimization elements for traffic lights and the simulation on SUMO has shown efficiency of the model \cite{13}. However, most of the works focused on single-agent controlling signals for an intersection.
\section{System Model }
The proposed system is shown in Fig. 1. The model for each agent is based on four items $\langle S, A, R, P\rangle$. Let $S$ is possible state space, and $s$ is a state ($s\in S$). $A$ is possible action space, and $a$ is an action ($a\in A$) and $R$ is reward space. Let $P$ is the transition probability function space from one state to next state. A series of consequent actions is policy $\pi$. Learning an optimal policy to maximize the cumulative expected reward is main goal. An agent at state $s$ takes an action $a$ to reach next state $s'$, gets the reward $r$.
\begin{figure}[t]
\centering
\includegraphics[width=1\linewidth]{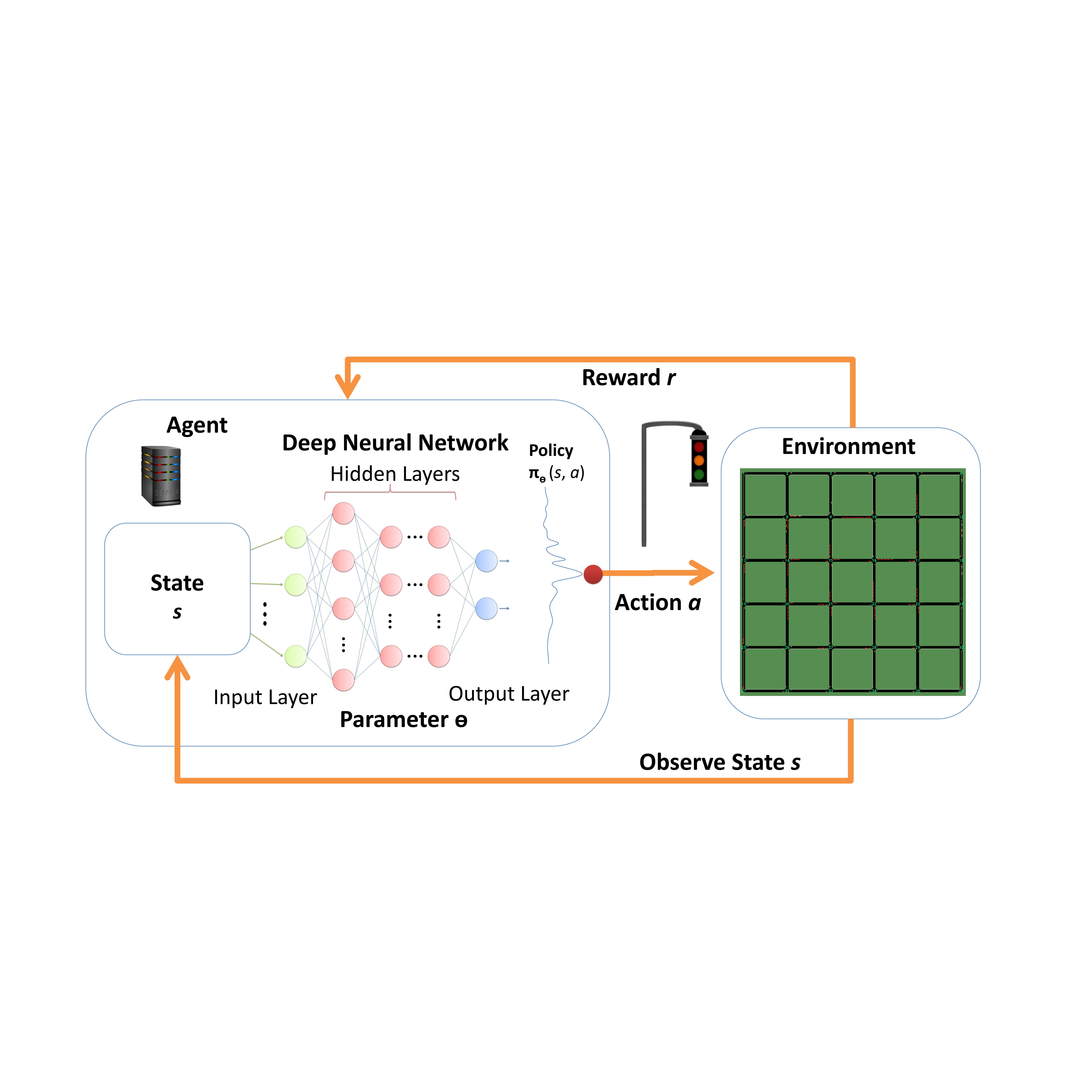}
\caption{Illustration of system under study}
\end{figure}
A four-tuple represents this situation as $\langle$$s$,$a$,$r$,$s'$$\rangle$. The state transition occurs at a discrete step $n$ in $\pi$. Let $Q(s,a)$ is cumulative reward function in future by executing an action $a$ at state $s$. Let $r_n$ is reward at $n$th step, and (1) gives $Q^{\pi}(s,a)$ for $\pi$:
\begin{equation}
\label{eq:emc}
Q^{\pi}(s,a)=E[\sum\limits_{k=0}^{{\infty}}{{\gamma ^k r_{n+k}|s_n=s,a_n=a,\pi }}] \
\end{equation}
The parameter $\gamma$ is a discount factor in $[0,1)$, decides how much importance should be given to recent and future rewards. The optimal policy $\pi^*$ can be acquired through several episodes in the learning process. Calculation of optimal $Q(s,a)$ is based on the optimal $Q$ values of the succeeding states represented by the Bellman optimality (2):
\begin{equation}
\label{eq:emc}
Q^{\pi^*}(s,a)=E_{s'}[r_n+\gamma \max_{a'}Q^{\pi^*}(s',a')|s,a] \
\end{equation}
It can be solved by dynamic programming keeping finite states for less computational burden. However, the $Q$ values can also be estimated by a function $\theta$ for larger number of states.
\subsection{Problem formulation}
The environment (road traffic) is shared by the agents. Let $D_a$ is total number of controlled lanes of an agent $a$. Where $a\subseteq A$ and $A =\{2,4,8\}$. The number of waiting cars is $w_{{i_a}}$ where ${i_a}$ is a lane. The reward is $r_{a_c}$ considering a case $c\subseteq C$ and $C=\{1,2,3,4\}$. Figure 2 shows the multi-gent traffic scenario under study. We aim to minimize average of $w_{{i_a}}$ in each $c$ and varying number of multi-agents as described in problem \text{(P)}:
\begin{equation}
\label{eq:emc}
\text{(P)}: \min_{\pi} E_{\pi}\left[  \sum\limits_{i_a=1}^{{{D_a}}} {{{w_{i_a}}}}\right]s.t. r_{a_c}\in\{0,1\},\forall C,\forall A\
\end{equation}
\begin{figure}[]
\centering
\includegraphics[width=1\linewidth]{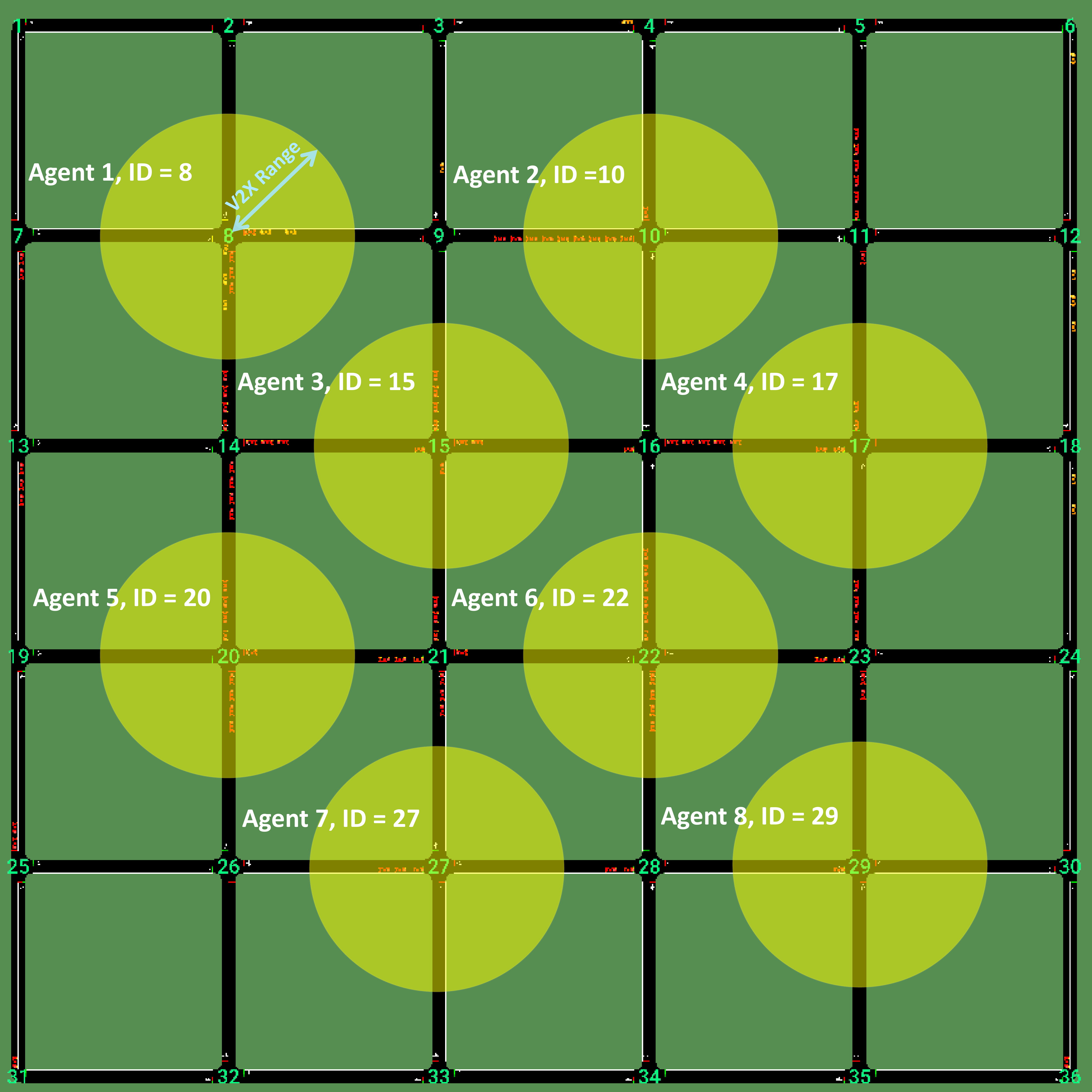}
\caption{Multi-agent traffic scenario}
\end{figure}
\subsubsection{States}
The states are number of vehicles on road for each lane of the traffic light agents. The number of vehicles are acquired from V2X communication in DSRC mode \cite{DSRC}. This reduces number of states to the controlled lanes for a multi-lane traffic intersection reducing computational burden. The length of road is defined as $l_r$. The state is a four-value vector $\langle l_0,l_1,l_2, l_3 \rangle$ such that each element represents number of vehicles respectively in lane 0, lane 1, lane 2 and lane3 (North, East, South, West).
\begin{figure}[]
\centering
\includegraphics[width=0.85\linewidth]{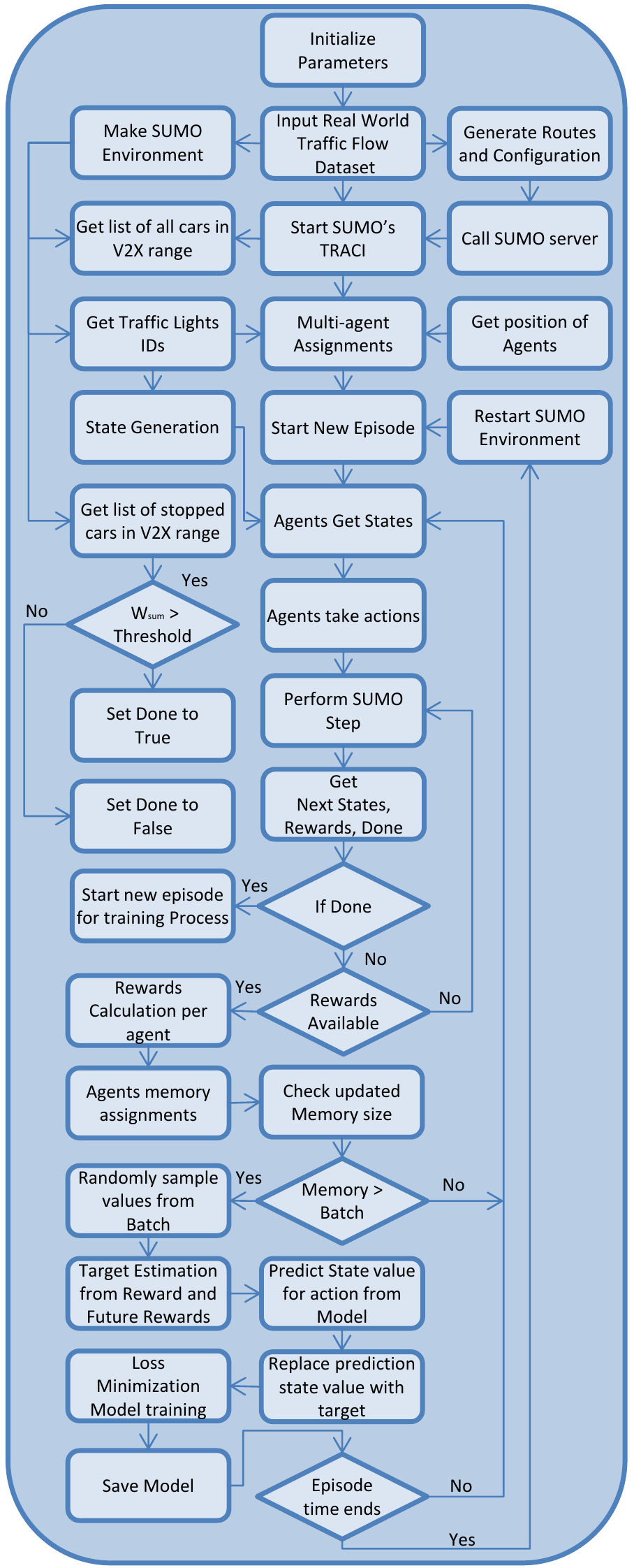}
\caption{Process flow of the multi-agent system under study}
\end{figure}
\subsubsection{Actions}
The actions space decides duration of each phase in the next cycle. The duration of two phase changes between two consecutive cycles is modeled as a high-dimension MDP. The phase remains constant during $D$ seconds. Let $\langle D1,D2,D3,D4\rangle$ are durations of 4 phases. The duration of one phase in the next cycle will be incremented by $D$ if the same action is chosen by an agent. Repeating an action will increase duration of the same phase. To investigate feasibility of actions, we assume that probability of phase transition for action $i$ to action $j$ is $P_{ij}$. Let $f_{ij(n)}$ is the probability that a chosen action starting from $i$ will go to $j$ for the first time after $n$ steps. An agent may take too long time to choose an unvisited action if there are no bounded conditions. The probability that an action $i$ will be chosen after action $j$ in one action-step is $f_{ij(1)}$ as in (4):
\begin{equation}
\label{eq:emc}
f_{ij}(1)=P_{ij}^{(1)}= P_{ij}\
\end{equation}
The first passage probability after $n$ action-steps can be generalized as (5):
\begin{equation}
\label{eq:emc}
f_{ij}(n)={P_{ij}^{(n)}} - f_{ij}{(1)}P_{jj}^{(n-1)} -...f_{ij}{(n-1)}P_{jj}^{(1)}\
\end{equation}
Let $q_{ij}$ be that probability that an agent at action $i$ will eventually take $j$ at least once in $m$ transitions then it will be the sum of all first passage probabilities (6):
\begin{equation}
\label{eq:emc}
q_{ij}(m)={f_{ij}{(1)}} + {f_{ij}{(2)}}+{f_{ij}{(3)}}+...+{f_{ij}{(m)}}\
\end{equation}
The feasible operation of the proposed system is possible if and only if $j$ is transient in nature which is shown by (7):
\begin{equation}
\label{eq:emc}
q_{ij}(\infty)=\sum \limits_{n=1}^{{\infty}} P_{jj}^{(n)}<0\
\end{equation}
 It can achieved by setting a threshold $W_{sum}$ for the maximum waiting number of vehicles.
%\begin{figure}[t]
%\centering
%\includegraphics[width=0.8\linewidth]{Figure1-state}
%\caption{A state diagram for the actions}
%\end{figure}
\subsubsection{Rewards}
Our study aims at decreasing $w_{{i_a}}$ in each lane. This proportionally reduces cumulative waiting time. In contrast to previous research work of single agent we argue that real-life scenarios consist of multiple traffic lights and the learning process of one agent may not be effective in reducing traffic congestion in the neighborhood.
The reason is that, once an agent performs good at an intersection, then the congestion at the connected roads of this agent will be reduced causing increase in traffic flow which will result in severe traffic jam in the neighboring intersections. To investigate this we consider four cases in $C$.
\begin{figure*}[t]
\centering
\includegraphics[width=0.8\linewidth]{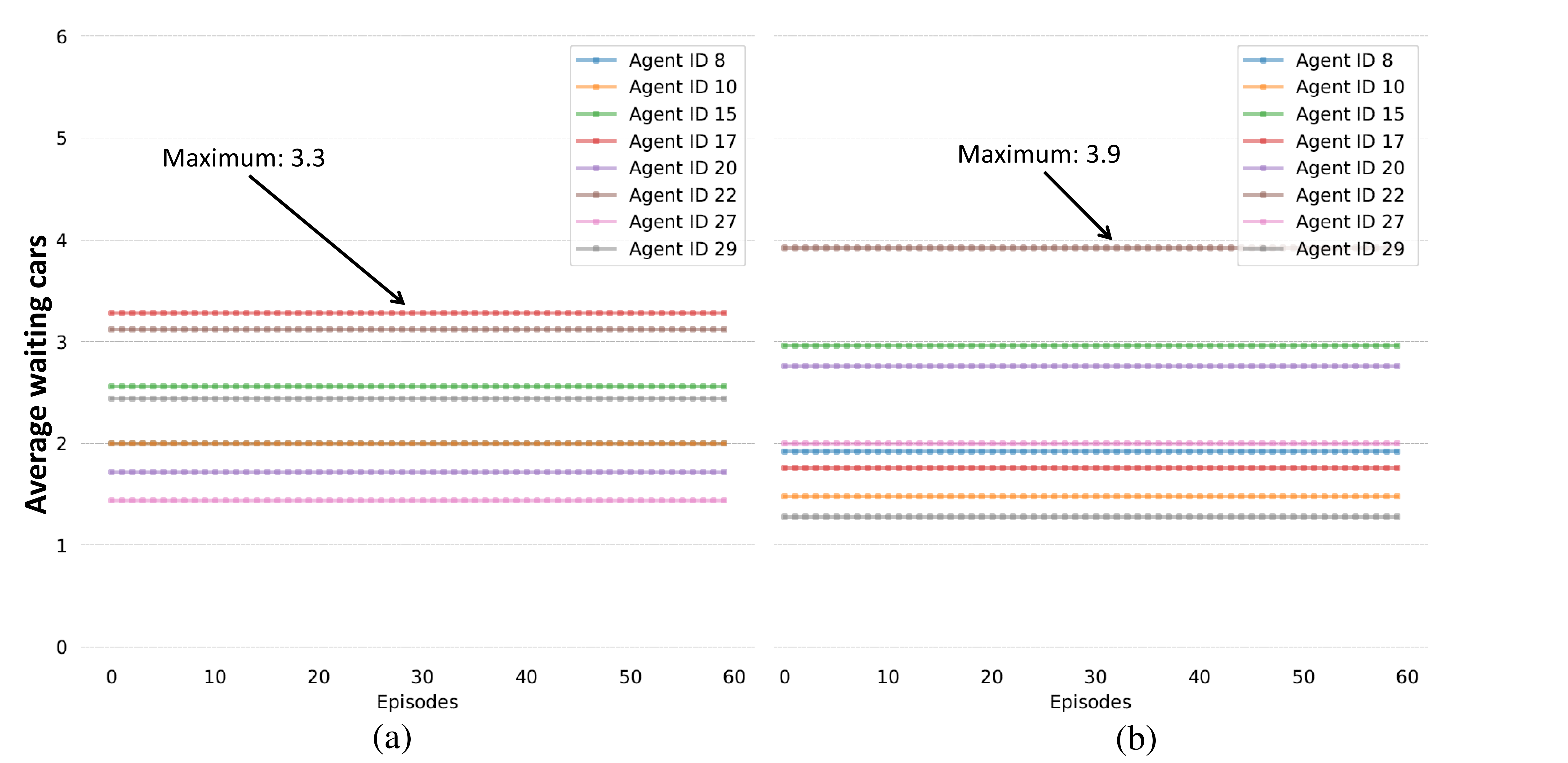}
\caption{Results for traditional periodic TLC. (a) Duration per phase is 30 s. (b) Duration per phase is 40 s.}
\end{figure*}
\subsubsection{Agents with unshared-rewards}
In first case the reward for an agent $a$, is accumulative waiting number of vehicles in its vicinity. Let $D_a$ is total number of lanes of agent $a$. The reward $r_{a_1}$, for case 1 is given by (8):
\begin{equation}
\label{eq:emc}
r_{a_1}=({{1+\sum\limits_{i_a=1}^{{D_a}}{w_{i_a}}}})^{-1}, \forall \begin{cases}{{\sum\limits_{i_a=1}^{{D_a}}{w_{i_a} }}}<W_{sum}
\end{cases}\
\end{equation}
The reward in the second case is accumulative waiting time of vehicles in an agent's vicinity. Let $w_{t,{i_a}}$ is total waiting time of all vehicles in ${i_a}$, (9) gives the reward $r_{a_2}$ for a step $t$,
\begin{equation}
\label{eq:emc}
r_{a_2}=({{1+\sum\limits_{i_a=1}^{{D_a}}{w_{t,{i_a}}}}})^{-1}\
\end{equation}
\subsubsection{Agents with shared-rewards}
The third case considers common aim for each agent. The actions of agents are selected to minimize the overall waiting number of cars of all agents. Therefore, reward for each agent is the accumulative waiting number of vehicles in its own as well as other agents vicinity. Let $A$ number of agents are selected for the common environment, $D_b$ is total number of agents except the agent $a$, the reward $r_{a_3}$ is given by (10):
\begin{equation}
\label{eq:emc}
r_{a_3}=({{1+\sum\limits_{i_a=1}^{{D_a}}{w_{i_a}}+\sum\limits_{b=1,b\neq a}^{{A}}\sum\limits_{i_{b}=1}^{{D_{b}}}{w_{i_{b}}}}})^{-1}\
\end{equation}
In fourth case we consider waiting time experienced by all agents as the shared reward for an agent. It means that $r_{a_3}$ is accumulative waiting time of vehicles in its own and other agents vicinity as expressed in (11):
\begin{equation}
\label{eq:emc}
r_{a_4}=({{1+  \sum\limits_{i_a=1}^{{D_a}}{w_{t,i_a}}+\sum\limits_{b=1,b\neq a}^{{A}}          \sum\limits_{i_{b}=1}^{{D_{b}}}{w_{t,i_{b}}}}})^{-1}\
\end{equation}
Variable $w_{t,{i_b}}$ is for total waiting time of all vehicles in ${i_b}$.
\subsection{Process flow of the proposed model}
Figure 3 shows process flow of the proposed multi-agent model. The initialization parameters and their values are in Table I. The proposed algorithm reads the dataset of traffic flow per lane and formulate the vehicle flow rate from real world domain to SUMO domain in terms of steps.

We choose this formulation to assign vehicle flow rate or arrival probability to each vehicle according to configuration 1 of ``Huawei 2019 Vehicle Scheduling Contest'' \cite{Hua}. The purpose of selecting this online dataset is to establish a standard comparison for the research. A total of 128 cars are used. The process of agent training starts by initializing the SUMO environment. The SUMO environment imitates the real world road network with traffic lights and vehicles. Each agent is responsible for its traffic light region. The V2X coverage is limited to $C_r$ meter radius of a circle around the traffic light intersection. The agent gets information of vehicles in each connected roads under the coverage area.

Each lane is given equal importance in the calculation of reward function. All agents take actions in a predefined cycle duration. The agents act upon their respective states either using experience replay or based on the random decision under exploration. The experience replay is used individually by each agent to minimize its loss. The loss is difference between target and prediction. A separate neural network acts as a function approximator known as Q-network $Q(s,a;\theta)$ with weights $\theta$. The Q-network is trained by minimizing the sequence of loss functions $L_i(\theta_i)$ which changes in each $i$th iteration is shown in Equation (12):
\begin{equation}
\label{eq:emc}
L_i(\theta_i)=E_{s,a\sim P(s,a)}[(Q_{target_{i}}(s,a)-Q(s,a;\theta_i))^2]\
\end{equation}
where $P(s,a)$ is the probability distribution over states and action sequences and $Q_{target_{i}}(s,a)$ for the $i$th iteration is given by the Equation (13):
\begin{equation}
\label{eq:emc}
Q_{target_{i}}(s,a)=E_{s'\sim \varepsilon}[r+\gamma \max_{a'}Q(s',a';\theta_{i-1})|s,a] \
\end{equation}
The weights of previous iteration $\theta_{i-1}$ are kept fixed during the optimization of loss function $L_i(\theta_i)$. The term $\varepsilon$ is SUMO environment. The aim is to predict the Q value for a given state and action, get the target value and replace predicted state for that action with the target value. The targets are dependent on the Q network weights.
Let $L_i = (r+\gamma \max_{a'}Q(s',a';\theta_{i-1})-Q(s,a;\theta_i)$ is a temporary variable. The differentiation of the loss function with respect to the weights gives gradient in the form of Equation (14):
\begin{equation}
\label{eq:emc}
\nabla_{\theta_i}L_i(\theta_i)=E_{s,a\sim P(s,a)}[L_i \nabla Q(s,a;\theta_i)]\
\end{equation}
\begin{figure*}[]
\centering
\includegraphics[width=0.9\linewidth]{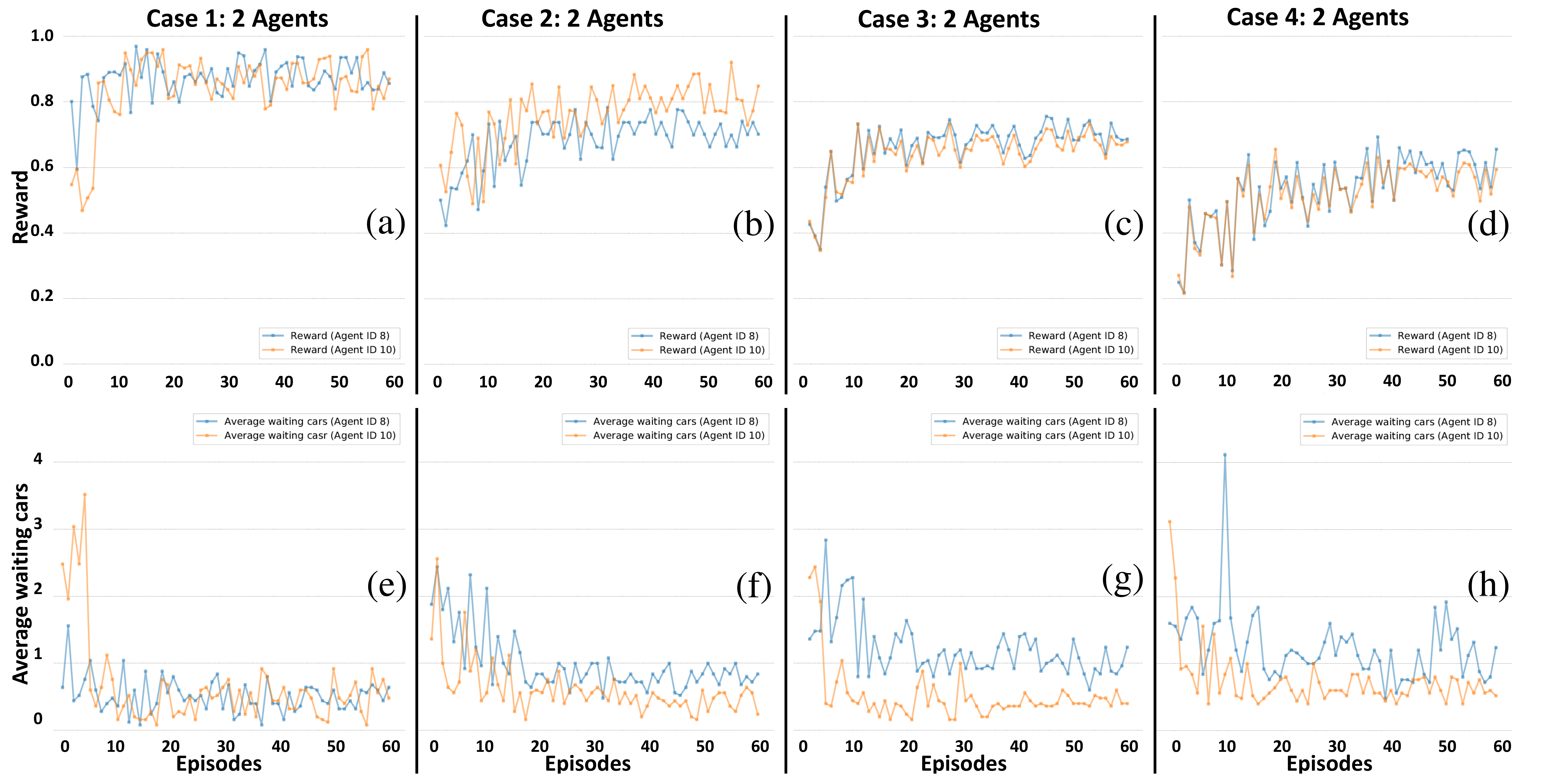}
\caption{Results with 2 agents. (a) Average of reward $r_{a_1}$. (b) Average of reward $r_{a_2}$. (c) Average of reward $r_{a_3}$. (d) Average of reward $r_{a_4}$. (e) Average waiting cars for case 1. (f) Average waiting cars for case 2. (g) Average waiting cars for case 3. (h) Average waiting cars for case 4.}
\end{figure*}
The proposed multi-agent DRL is a model-free approach. It solves the tasks by directly using the samples from SUMO environment and does not require the estimate of $\varepsilon$. Each agent learns its greedy strategy $\max_{a}Q(s,a;\theta)$ and follows a behaviour distribution ensuring adequate state space exploration. The learning behaviour distribution follows $\epsilon$-greedy strategy, which means 1-$\epsilon$ is selected for the exploitation and random action is selected with the probability $\epsilon$.
\begin{table}[b]
\scriptsize
\centering
\caption{Parameters for model evaluation}
\label{my-label}
\begin{tabular}{ll}
\hline
\textbf{Parameter} & \textbf{ Value}                                                 \\ \hline
Learning Rate       & 0.001               \\
Memory Size        & 10000                                      \\
Epsilon initial         & .95                                    \\
Epsilon final       &0.01                                    \\
Epsilon Decay rate & 0.001\\
Minibatch Size &32 \\
Discount Factor $\gamma$ & 0.95\\
Target network activation ReLU & beta 0.01\\
Loss metric & MSE\\
Optimizer & Adam\\
V2X circular coverage area & 45,216 $m^2$\\
Episodes & 60 \\
Fully connected Hidden Layers & 3 \\
Nodes per each hidden layer & 24, 24, 24 \\
\hline
\end{tabular}
\end{table}
\section{Evaluation}

The scenario is a 6$\times$6 intersections scenario with 2, 4, 8 multi-agents are selected as shown in Fig. 3. There are a total of 128 vehicles that randomly enter the scenario from various intersections. The parameters used for training the deep neural network are in Table I. Figure 4 shows the average waiting cars at 8 intersections for the scenario. In Fig. 4(a) we have kept 30 s duration for each phase. Similarly in Fig. 4(b) the duration is 40 s. It is noted that Agent IDs are reference intersection ID. The agents are not taking actions in the Fig. 5. It is observed that traffic at intersection marked with ID 22 has a higher (greater than 3) average number of waiting cars in both 30 s and 40 s cases. All intersections have more than 1 average waiting cars during all episodes.

We divide the parameters for evaluation as: model parameters and traffic parameters. The model is trained iteratively in each episode for 500 s. The reward and $w_{{i_a}}$ are aggregated in each episode. The performance of agents are compared for $r_{a_{1}}$, $r_{a_{2}}$, $r_{a_{3}}$ and $r_{a_{4}}$. The rewards in Fig. 5(a) are for agents with IDs 8 and 10 and evaluated with $r_{a_{1}}$. Similarly Fig. 5(b) shows corresponding rewards evaluated by using $r_{a_{2}}$. The shared-rewards $r_{a_{3}}$ and $r_{a_{4}}$ are respectively presented in Fig. 5(c) and Fig. 5(d). The effects of using different reward functions are shown below for each case in Fig. 5(e)-(h). It is observed that in case 1 both agents have performed better in maximizing their rewards and minimizing the number of waiting cars in their connected lanes. Agent 10 performed better than agent 8 for the cases 2, 3 and 4. The reward for the agent 10 is higher than agent 8 and the corresponding results of average waiting cars also show less average waiting cars for the agent 10 as compared to agent 8.

%
%\begin{algorithm}[]
%\caption{Multi-agent Traffic lights Control}
%\KwIn{Replay memory size M, batch size B, total episodes N$_{epi}$, total steps in an episode N$_{steps}$, number of agents N$_{agents}$.}
%
%Initialize SUMO environment based on route files generating from real traffic flow data file.Initialize L$_{threshold}$ as the maximum of vehicles during an episode in the V2X range of an agent.\\
%Start SUMO's TRACI\\
%\For{n$_{epi}$ = 1 : N$_{epi}$}{
%    Initialize states with the starting scenario for all agents\;
%    \For{t =1 : N$_{steps}$}{
%        \For{i =1 : N$_{agents}$}{
%            Choose an action a according to the e-greedy;\\
%            Perform SUMO step and get list of cars' id L in range as states of agent$_i$.\\
%            \eIf{len(L)\textgreater L$_{threshold}$ or t == N$_{steps}$}
%            {Break to the next episode\;}
%            {Observe reward r and new state s' and append them into memory. }
%            \If{length(M)\textgreater B}
%            {Select B samples from memory and estimate target from reward and future rewards\;Predict State value for action from model\;Minimize loss after replacing prediction state value with target\;Train model\;Save model.}
%
%        }
%    }
%
%}
%\end{algorithm}

Figure 6 shows results of 4 agents with IDs 8, 10, 15, and 17 by experimenting the same four cases using the similar scenario. It is observed in Fig. 6(a) that agent 10 performs better than other agents as the average number of waiting cars under agent 10 are less than others. In Fig. 6(b) all agents try to perform better by fluctuating their rewards. These fluctuations are also reflected in the number of waiting cars as shown in Fig. 6(f). Interesting similar rewards are observed in the shared-rewards cases (3,4) in Fig 6(c) and Fig 6(d).
\begin{figure*}[]
\centering
\includegraphics[width=0.9\linewidth]{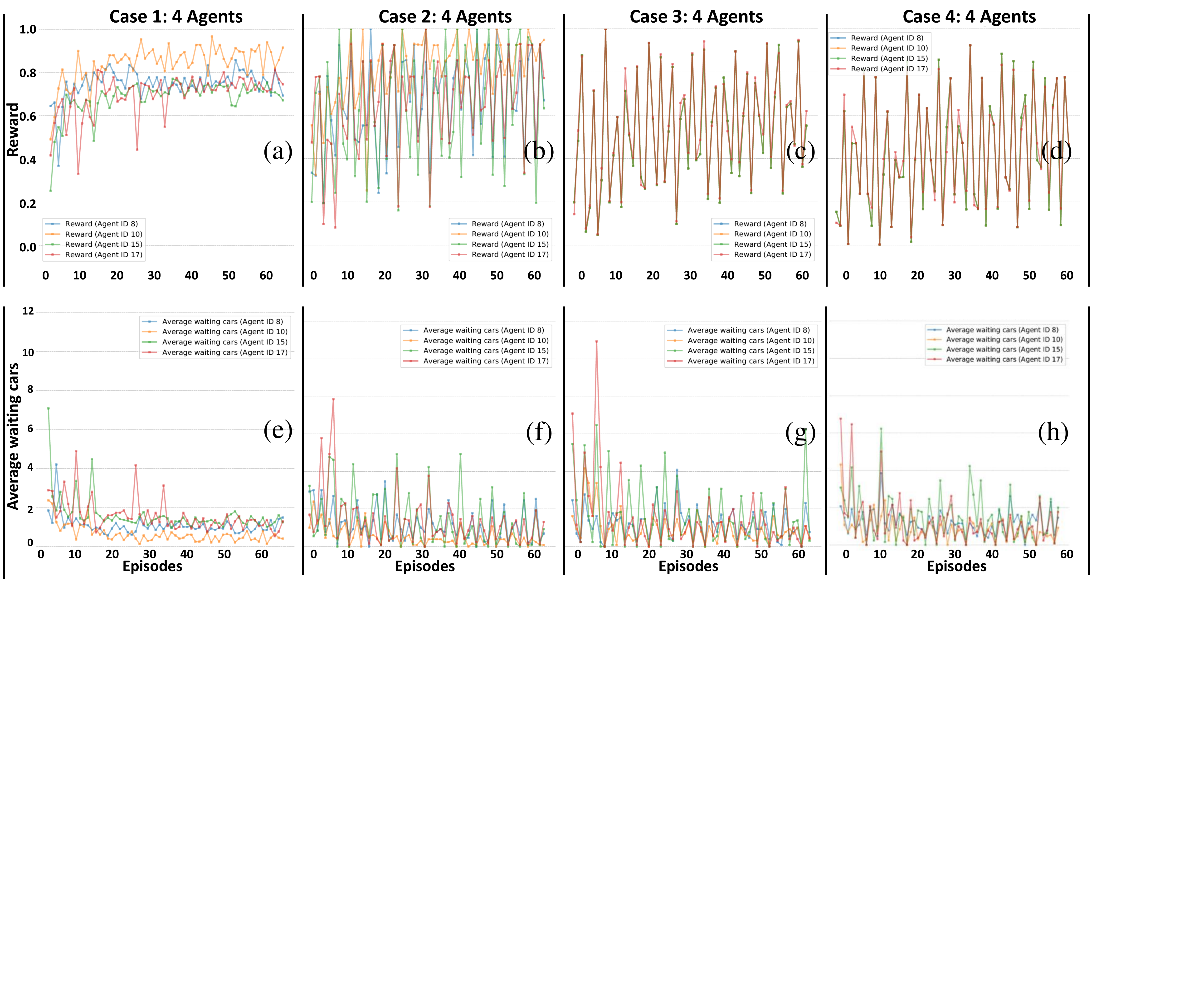}
\caption{Results with 4 agents. (a) Average of reward $r_{a_1}$. (b) Average of reward $r_{a_2}$. (c) Average of reward $r_{a_3}$. (d) Average of reward $r_{a_4}$. (e) Average waiting cars for case 1. (f) Average waiting cars for case 2. (g) Average waiting cars for case 3. (h) Average waiting cars for case 4.}
\end{figure*}
\begin{figure*}[b]
\centering
\includegraphics[width=0.9\linewidth]{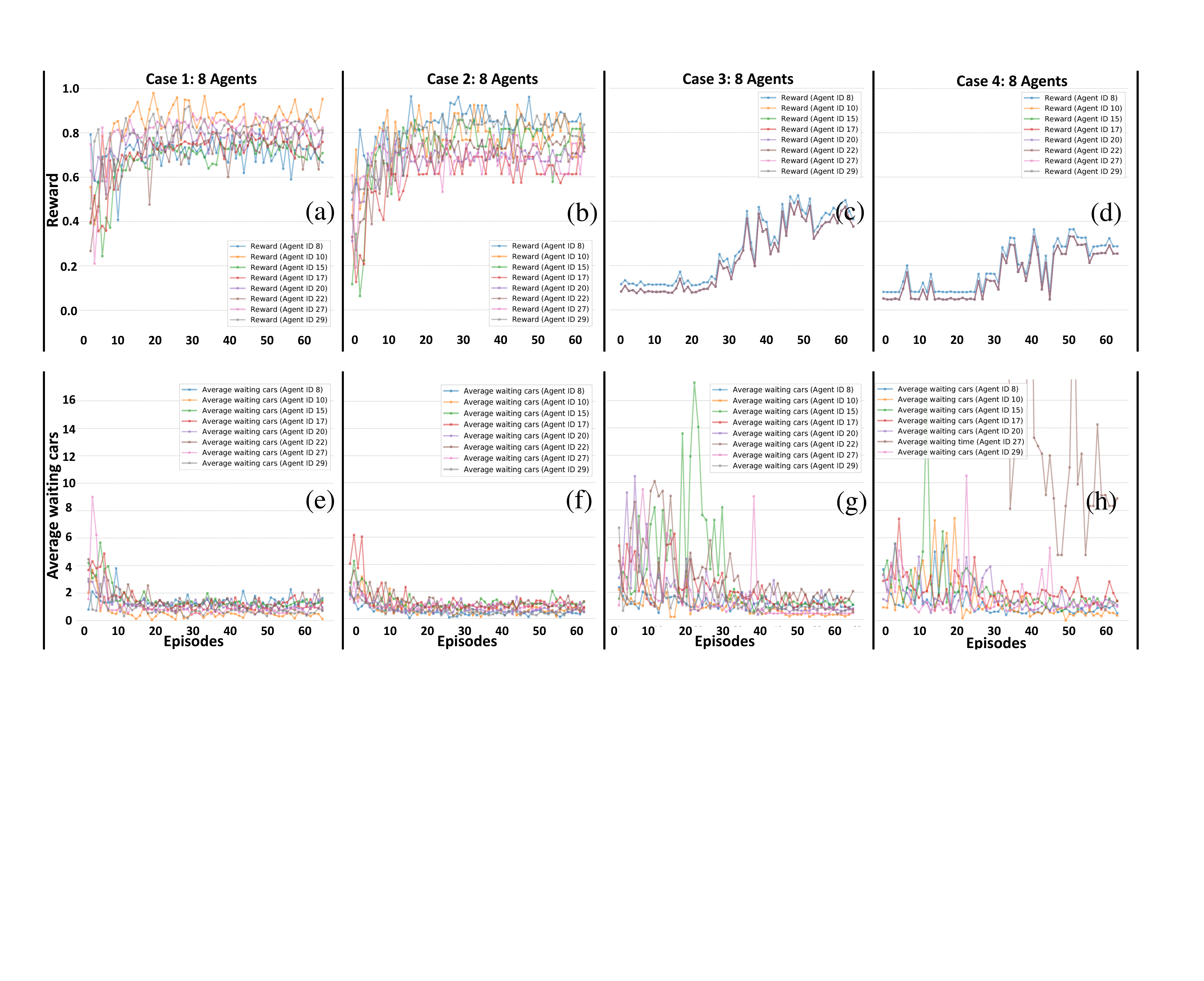}
\caption{Results with 8 agents. (a) Average of reward $r_{a_1}$. (b) Average of reward $r_{a_2}$. (c) Average of reward $r_{a_3}$. (d) Average of reward $r_{a_4}$. (e) Average waiting cars for case 1. (f) Average waiting cars for case 2. (g) Average waiting cars for case 3. (h) Average waiting cars for case 4.}
\end{figure*}
Average number of waiting cars are also similar as shown in Fig. 6(g) and Fig. 6(h). The results for the 8 agents are shown in Fig. 7 with IDs 8,10,15, 17, 20, 22, 27, and 29. The case 1 outperforms other cases in Fig 7(a). All agents tried to maximize their rewards showing better results in Fig. 7(a)-(b) and Fig. 7(e)-(f). Peak average waiting cars in Fig. 7 (f) is 2.1 as compared to 3.9 (see Fig. 4(b)) after 15 episodes, which means a decrease of 41.5 $\%$. On the other hand all rewards become same for the agents as expected. However, the shared-reward schemes performed poorly as compare to unshared-rewards. The reason behind this low performance is failure to achieve better actions that could maximize the shared reward. The actions of one agent produces the environment conditions that could negatively disturb the rewards for other agents. Agents 8, 10, 17, and 29 relatively performed better even with reduced shared rewards.
\section{Conclusion}
In this article we have presented the performance of deep reinforcement learning under multi-agent V2X driven traffic control system. We have observed that multi-agents with individual rewards considering waiting number of cars is a better choice as compared to the average waiting time. On the other hand shared-rewards based cases do not perform better. Shared-rewards make the situation more competitive. This competition should be further investigated using other techniques of deep reinforcement learning. We have also observed that for larger number of agents, the reward based on waiting time is the better choice.

% conference papers do not normally have an appendix

%% use section* for acknowledgment
%\section*{Acknowledgment}
%
%
%The authors would like to thank...

% trigger a \newpage just before the given reference
% number - used to balance the columns on the last page
% adjust value as needed - may need to be readjusted if
% the document is modified later
%\IEEEtriggeratref{8}
% The "triggered" command can be changed if desired:
%\IEEEtriggercmd{\enlargethispage{-5in}}

% references section

% can use a bibliography generated by BibTeX as a .bbl file
% BibTeX documentation can be easily obtained at:
% http://www.ctan.org/tex-archive/biblio/bibtex/contrib/doc/
% The IEEEtran BibTeX style support page is at:
% http://www.michaelshell.org/tex/ieeetran/bibtex/
%\bibliographystyle{IEEEtran}
% argument is your BibTeX string definitions and bibliography database(s)
%\bibliography{IEEEabrv,../bib/paper}
%
% <OR> manually copy in the resultant .bbl file
% set second argument of \begin to the number of references
% (used to reserve space for the reference number labels box)

% that's all folks
\end{document}